\documentclass[letterpaper]{article} 
\usepackage{aaai23}  
\usepackage{times}  
\usepackage{helvet}  
\usepackage{courier}  
\usepackage[hyphens]{url}  
\usepackage{graphicx} 
\usepackage{natbib}  
\usepackage{caption} 
\usepackage{algorithm}
\usepackage{algorithmic}
\usepackage{multirow}
\usepackage{subfigure}
\usepackage{booktabs}
\usepackage{amsfonts}
\usepackage{graphicx}
\usepackage{arydshln}
\usepackage{amsmath}

\usepackage{newfloat}
\usepackage{listings}
\DeclareCaptionStyle{ruled}{labelfont=normalfont,labelsep=colon,strut=off} 
\floatstyle{ruled}
\newfloat{listing}{tb}{lst}{}
\floatname{listing}{Listing}
\title{Adversarial Self-Attention for Language Understanding}
\author{
    Hongqiu Wu\textsuperscript{\rm 1,2},
    Ruixue Ding\textsuperscript{\rm 4},
    Hai Zhao\textsuperscript{\rm 1,2,\thanks{Corresponding author; This paper was partially supported by Key Projects of National Natural Science Foundation of China (U1836222 and 61733011).}},
    Pengjun Xie\textsuperscript{\rm 4},
    Fei Huang\textsuperscript{\rm 4},
    Min Zhang\textsuperscript{\rm 3}\\
}
\affiliations{
    \textsuperscript{\rm 1}Department of Computer Science and Engineering, Shanghai Jiao Tong University\\
    \textsuperscript{\rm 2}Key Laboratory of Shanghai Education Commission for Intelligent Interaction\\
        and Cognitive Engineering, Shanghai Jiao Tong University\\
    \textsuperscript{\rm 3}School of Computer Science and Technology, Soochow University\\
    \textsuperscript{\rm 4}Damo Academy, Alibaba Group\\

    wuhongqiu@sjtu.edu.cn, zhaohai@cs.sjtu.edu.cn,\\
    \{ada.drx,chengchen.xpj,f.huang\}@alibaba-inc.com, minzhang@suda.edu.cn
}

\usepackage{bibentry}

\begin{document}

\maketitle

\begin{abstract}
Deep neural models (e.g. Transformer) naturally learn spurious features, which create a ``shortcut'' between the labels and inputs, thus impairing the generalization and robustness. This paper advances the self-attention mechanism to its robust variant for Transformer-based pre-trained language models (e.g. BERT). We propose \textit{Adversarial Self-Attention} mechanism (ASA), which adversarially biases the attentions to effectively suppress the model reliance on features (e.g. specific keywords) and encourage its exploration of broader semantics. We conduct a comprehensive evaluation across a wide range of tasks for both pre-training and fine-tuning stages. For pre-training, ASA unfolds remarkable performance gains compared to naive training for longer steps. For fine-tuning, ASA-empowered models outweigh naive models by a large margin considering both generalization and robustness\footnote{\url{https://github.com/gingasan/adversarialSA}}.
\end{abstract}

\section{Introduction}

The emerging pre-trained language models (PrLMs) like BERT \citep{DBLP:conf/naacl/DevlinCLT19} have become the backbone of nowadays natural language processing (NLP) systems. It seems to reach the bottleneck in the recent NLP community. This paper rethinks the dilemma from the perspective of self-attention mechanism (SA) \citep{DBLP:conf/nips/VaswaniSPUJGKP17}, which is broadly chosen as the fundamental architecture of PrLMs and proposes \textit{Adversarial Self-Attention} mechanism (ASA).

\begin{figure}
\centering
\subfigure[SA]{
\includegraphics[width=0.22\textwidth]{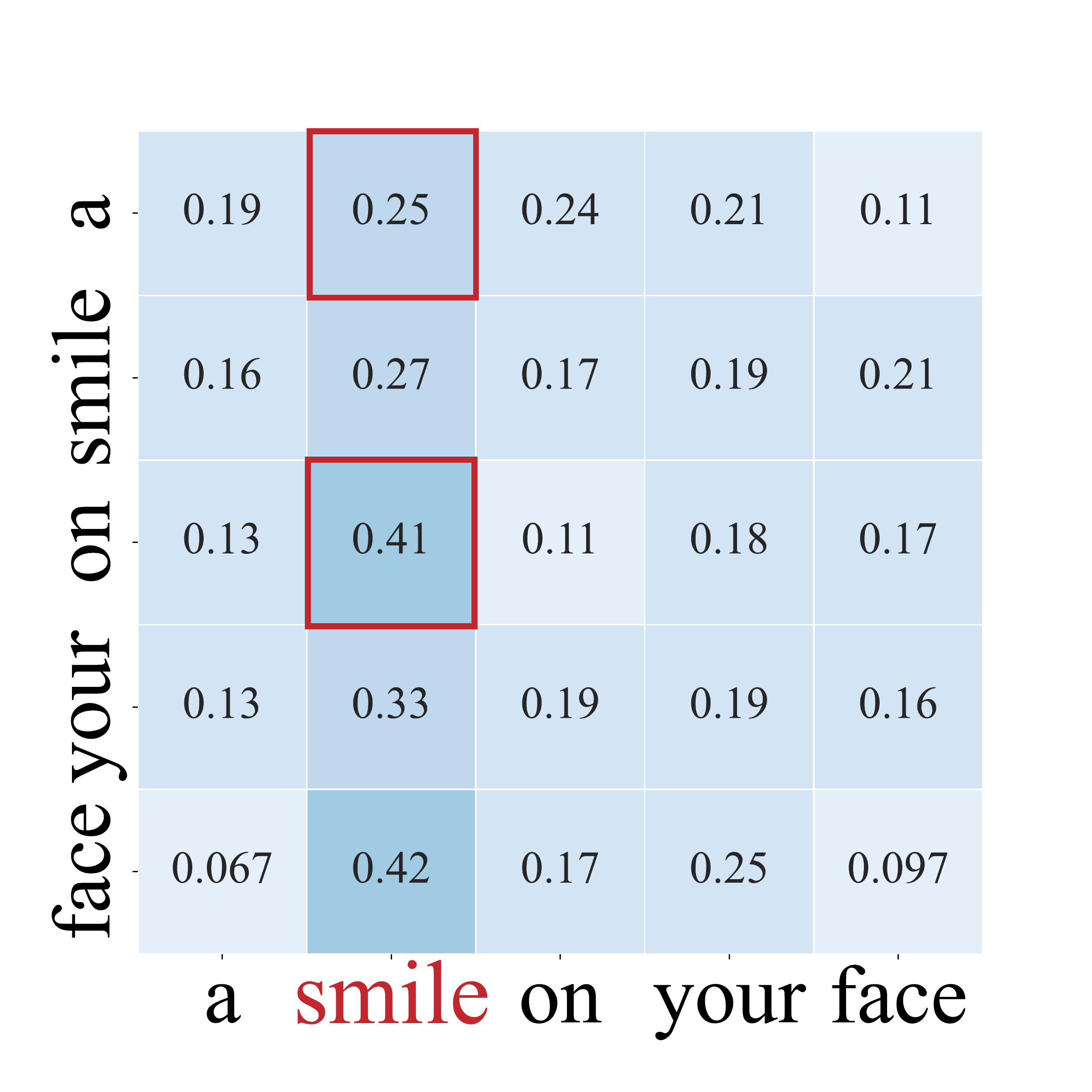}
}
\subfigure[ASA]{
\includegraphics[width=0.22\textwidth]{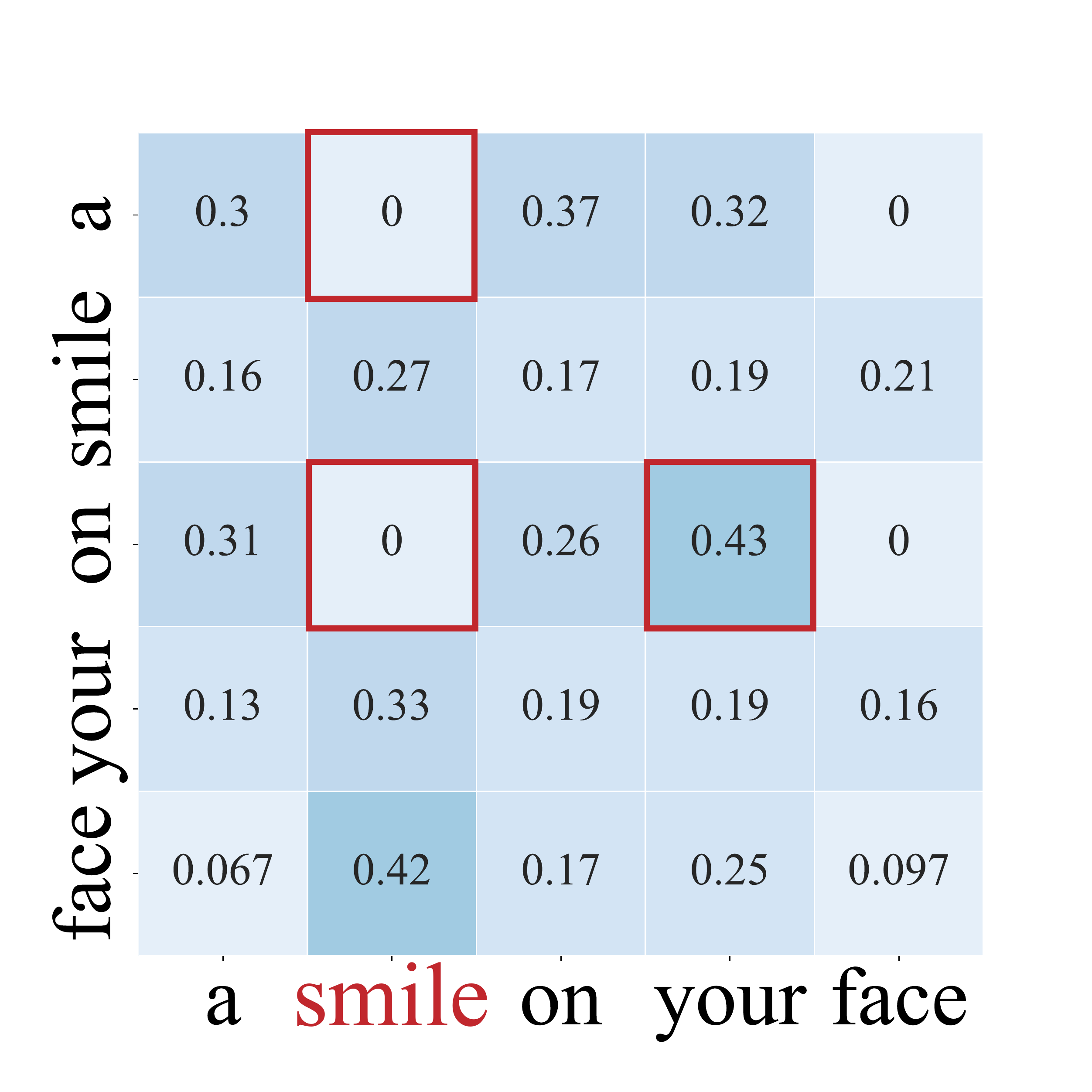}
}
\caption{Empirical attention maps from SA and ASA of the first BERT layer. The text is \textit{a smile on your face} from SST-2 dataset. The sum of horizontal scores is equal to 1. We highlight some units receiving strong attention diversions.}
\label{f5}
\end{figure}

\begin{figure*}
\centering
\includegraphics[width=0.89\textwidth]{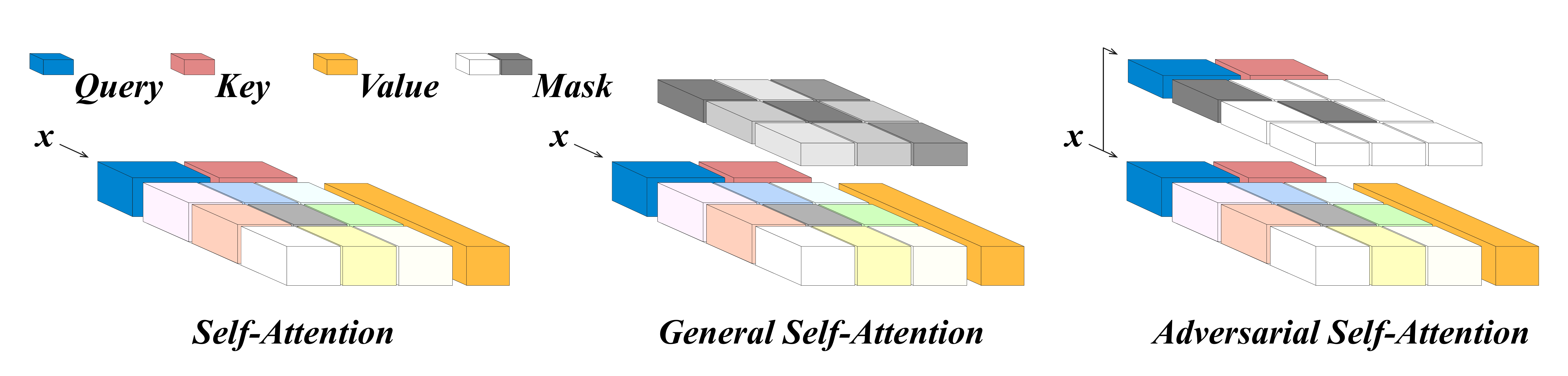}
\caption{Sketches from self-attention to adversarial self-attention. In self-attention, the attention matrix (the middle colored blocks) is straightforwardly from the query and key components. In general self-attention, the attention matrix is considered to be imposed with another matrix such that its distribution can be biased. In adversarial self-attention, the biasing matrix is learned from the input and is a binary mask (black and white).}
\label{f1}
\end{figure*}

A large body of empirical evidence \citep{DBLP:conf/icml/ShiGRXLLK21,DBLP:conf/acl/YouSI20,DBLP:conf/aaai/0001WZDZ020,DBLP:conf/acl/WuZZ21} indicates that self-attention can benefit from allowing bias, where researchers impose certain priorities (e.g. masking, smoothing) on the original attention structure to compel the model to pay attention to those ``proper'' tokens. We attribute such phenomena to the nature of deep neural models, which lean to exploit the potential correlations between inputs and labels, even spurious features \citep{DBLP:conf/icml/SrivastavaHL20}. It is harmful for generalizing on test data if the model learns to attend to spurious tokens. Priorly keeping the model form them can hopefully address this. However, crafting priorities is limited to task-specific knowledge and by no means lets the model training be an end-to-end process, whereas generating and storing that knowledge can be even more troublesome.

The idea of ASA is to adversarially bias the self-attention to effectively suppress the model reliance on specific features (e.g. keywords). The biased structures can be learned by maximizing the empirical training risk, which automates the process of crafting specific prior knowledge. Additionally, those biased structures are derived from the input data itself, which sets ASA apart from conventional adversarial training. It is a kind of Meta-Learning \citep{DBLP:books/sp/98/ThrunP98}. The learned structures serve as the ``meta-knowledge" that facilitates the self-attention process.

We showcase a concrete example in Figure \ref{f5}. For the word \textit{on}, it never attends to \textit{smile} after being attacked in ASA but strongly attends to \textit{your}. It can be found that \textit{smile} serves as a keyword within the whole sentence, suggesting a positive emotion. The model can predict the right answer based on this single word even without knowing the others. Thus, ASA tries to weaken it and let those non-keywords receive more attention. However, in a well-connected structure, the masked linguistic clues can be inferred from their surroundings. ASA prevents the model from shortcut predictions but urges it to learn from contaminated clues, which thus improves the generalization ability.

Another issue of concern is that adversarial training typically results in greater training overhead. In this paper, we design a simple and effective ASA implementation to obtain credible adversarial structures with no training overhead.

This paper is organized as follows. $\S$ 2 summarizes the preliminaries of adversarial training and presents a general form of self-attention. $\S$ 3 elaborates the methodology of the proposed ASA. $\S$ 4 reports the empirical results. $\S$ 5 compares ASA with naive masking as well as conventional adversarial training upon the performance and efficiency. $\S$ 6 takes a closer look at how ASA works.

\section{Preliminary}
\label{s2}

In this section, we lay out the background of adversarial training and self-attention mechanism. We begin with a number of notations. Let $ \mathbf x $ denote the input features, typically a sequence of token indices or embeddings in NLP, and $ y $ denote the ground truth. Given a model parameterized by $ \theta $, the model prediction can be thus written as $ p(y|\mathbf x,\theta) $.

\subsection{Adversarial Training}

Adversarial training (AT) \citep{DBLP:journals/corr/GoodfellowSS14} encourages model robustness by pushing the perturbed model prediction to $ y $:
\begin{equation}
\mathop{\min}_\theta \mathcal D \left[y,p(y|\mathbf x+\delta^*,\theta)\right]
\label{e1}
\end{equation}
where $ \mathcal D [\cdot] $ refers to KL divergence and $ p(y|\mathbf x+\delta^*,\theta) $ refers to the perturbed model prediction under an adversarial perturbation $ \delta^* $. In this paper, we apply the idea of virtual adversarial training (VAT) \citep{DBLP:journals/pami/MiyatoMKI19}, where $ y $ is smoothed by the model prediction $ p(y|\mathbf x,\theta) $. However, the smoothness can hold only if the amount of training samples is sufficiently large so that $ p(y|\mathbf x,\theta) $ can be so close to $ y $. This assumption is tenable for PrLMs. Even when fine-tuning on limited samples, the model can not predict so badly in the support of large-scale pre-trained weights.

The adversarial perturbation $ \delta^* $ is defined to maximize the empirical risk of training:
\begin{equation}
\delta^* = \mathop{\arg\max}_{\delta;\parallel \delta \parallel \le \epsilon} \mathcal D \left[p(y|\mathbf x,\theta),p(y|\mathbf x+\delta,\theta)\right]
\label{e2}
\end{equation}
where $ \parallel \delta \parallel \le \epsilon $ refers to the decision boundary restricting the adversary $ \delta $. We expect that $ \delta^* $ is so minor but greatly fools the model. Eq. \ref{e1} and Eq. \ref{e2} make an adversarial ``game'', in which the adversary seeks to find out the vulnerability while the model is trained to overcome the malicious attack.

\subsection{General Self-Attention}

Standard self-attention (SA, or vanilla SA) \citep{DBLP:conf/nips/VaswaniSPUJGKP17} can be formulated as:
\begin{equation}
{\rm SA}(Q,K,V)={\rm Softmax}\left(\frac{Q \cdot K^{T}}{\sqrt{d}}\right) \cdot V
\label{e3}
\end{equation}
where $ Q $, $ K $, and $ V $ refer to the query, key, and value components respectively, and $ \sqrt{d} $ is a scaling factor. In this paper, we define the pair-wise matrix $ {\rm Softmax}\left(\frac{Q \cdot K^{T}}{\sqrt{d}}\right) $ as the \textit{attention topology} $ \mathcal T(Q,K) $. In such a topology, each unit refers to the attention score between a token pair, and every single token is allowed to attend to all the others (including itself). The model learns such a topology so that it can focus on particular pieces of text.

However, empirical results show that manually biasing this process (i.e. determining how each token can attend to the others) can lead to better SA convergence and generalization, e.g. enforcing sparsity \citep{DBLP:conf/icml/ShiGRXLLK21}, strengthening local correlations \citep{DBLP:conf/acl/YouSI20}, incorporating structural clues \citep{DBLP:conf/acl/WuZZ21}. Basically, they smooth the output distribution around the attention structure with a certain priority $ \mu $. We call $ \mu $ the \textit{structure bias}. Therefore, it leads us to a more general form of self-attention:
\begin{equation}
{\rm SA}(Q,K,V,\mu)=\mathcal T(Q,K,\mu) \cdot V
\label{e4}
\end{equation}
where $ \mathcal T(Q,K,\mu) $ is the biased attention topology $ {\rm Softmax}\left(\frac{Q \cdot K^{T}}{\sqrt{d}}+\mu\right) $. In standard SA, $ \mu $ equals an all-equivalent matrix (all elements are equivalent). That means the attentions on all token pairs are unbiased.

The general form in Eq. \ref{e4} indicates that we are able to manipulate the way the attentions unfold between tokens via overlapping a specific structure bias $ \mu $. The corresponding sketches are in Figure \ref{t2}. We focus on the masking case, which is commonly used to mask out the padding positions, where $ \mu $ refers to a binary matrix with elements in $ \{0, -\infty\} $. When an element equals $ -\infty $, the attention score of that unit is off ($ =0 $). Note that the mask here is different from that in dropout \citep{DBLP:journals/jmlr/SrivastavaHKSS14} since the masked units will not be discarded but will be redistributed to other units.

\section{Adversarial Self-Attention Mechanism}
\label{s3}

This section presents the details of our proposed \textit{Adversarial Self-Attention} mechanism (ASA).

\subsection{Definition}
\label{asa}

The idea of ASA is to mask out those attention units to which the model is most vulnerable. Specifically, ASA can be regarded as an instance of general self-attention with an adversarial structure bias $ \mu^* $. However, those vulnerable units vary from inputs. Thus, we let $ \mu^* $ be a function of $ \mathbf x $, denoted as $ \mu_{\mathbf x}^* $. ASA can be eventually formulated as:  
\begin{equation}
{\rm ASA}(Q,K,V,\mu_{\mathbf x}^*)=\mathcal T(Q,K,\mu_{\mathbf x}^*) \cdot V
\label{e5}
\end{equation}
where $ \mu_{\mathbf x} $ is parameterized by $ \eta $ namely $ \mu_{\mathbf x}^*=\mu(\mathbf x,\eta^*) $. We also call $ \mu_{\mathbf x} $ the ``adversary'' in the following. Eq. \ref{e5} indicates that $ \mu_{\mathbf x}^* $ acts as ``meta-knowledge'' learned from the input data itself, which sets ASA apart from other variants where the bias is predefined based on a certain priority.

\subsection{Optimization}

Similar to adversarial training, the model is trained to minimize the following divergence:
\begin{equation}
\mathop{\min}_\theta \mathcal D \left[p(y|\mathbf x,\theta),p(y|\mathbf x,\mu(\mathbf x,\eta^*),\theta)\right]
\label{e6}
\end{equation}
where $ p(y|\mathbf x,\mu(\mathbf x,\eta^*),\theta) $ refers to the model prediction under the adversarial structure bias. We evaluate $ \eta^* $ by maximizing the empirical risk:
\begin{equation}
\eta^* = \mathop{\arg\max}_{\eta;\Vert \mu(\mathbf x,\eta) \Vert \le \epsilon} \mathcal D \left[p(y|\mathbf x,\theta),p(y|\mathbf x,\mu(\mathbf x,\eta),\theta)\right]
\label{e7}
\end{equation}
where $ \Vert \mu(\mathbf x,\eta) \Vert \le \epsilon $ refers to the new decision boundary for $ \eta $. The design of this constraint is necessary for keeping ASA from hurting model training.

Generally, researchers use $ L_2 $ or $ L_\infty $ norm to make the constraint in adversarial training. Considering that $ \mu(\mathbf x,\eta) $ is in form of a binary mask, it is more reasonable to constrain it by limiting the proportion of the masked units, which comes to $ L_0 $ or $ L_1 $ norm (since $ \mu(\mathbf x,\eta) $ is binary, they are the same), namely $ \Vert \mu(\mathbf x,\eta) \Vert_1 \le \epsilon $. The question is that it is cumbersome to heuristically develop the value of $ \epsilon $.  As an alternative, we transform the problem with a hard constraint into an unconstrained one with a penalty:
\begin{equation}
\begin{aligned}
\eta^* = &\mathop{\arg\max}_\eta \: \ \mathcal D \left[p(y|\mathbf x,\theta),p(y|\mathbf x,\mu(\mathbf x,\eta),\theta)\right]\\
       &+ \tau \Vert \mu(\mathbf x,\eta) \Vert_1
\end{aligned}
\label{e8}
\end{equation}
where we use a temperature coefficient $ \tau $ to control the intensity of the adversary. Eq. \ref{e8} indicates that the adversary needs to maximize the training risk and at the same time mask the least number of units as possible. Our experiments show that it is much easier to adjust $ \tau $ than to adjust $ \epsilon $ as in adversarial training. We find good performances when $ \tau=0.1\sim0.3 $.

Eventually, we generalize Eq. \ref{e8} to a model with $ n $ self-attention layers (e.g. BERT). Let $ \mu(\mathbf x,\eta)=\{\mu(\mathbf x,\eta)^1,\cdots,\mu(\mathbf x,\eta)^n\} $, where $ \mu(\mathbf x,\eta)^i $ refers to the adversary for the $ i^{\rm th} $ layer. The penalty term thus becomes the summation $ \Vert \mu(\mathbf x,\eta) \Vert_1=\sum_{i=1}^{n} \Vert \mu(\mathbf x,\eta)^i \Vert_1 $.

\subsection{Fast Implementation}
\label{fastasa}

\begin{figure}
\centering
\includegraphics[width=0.43\textwidth]{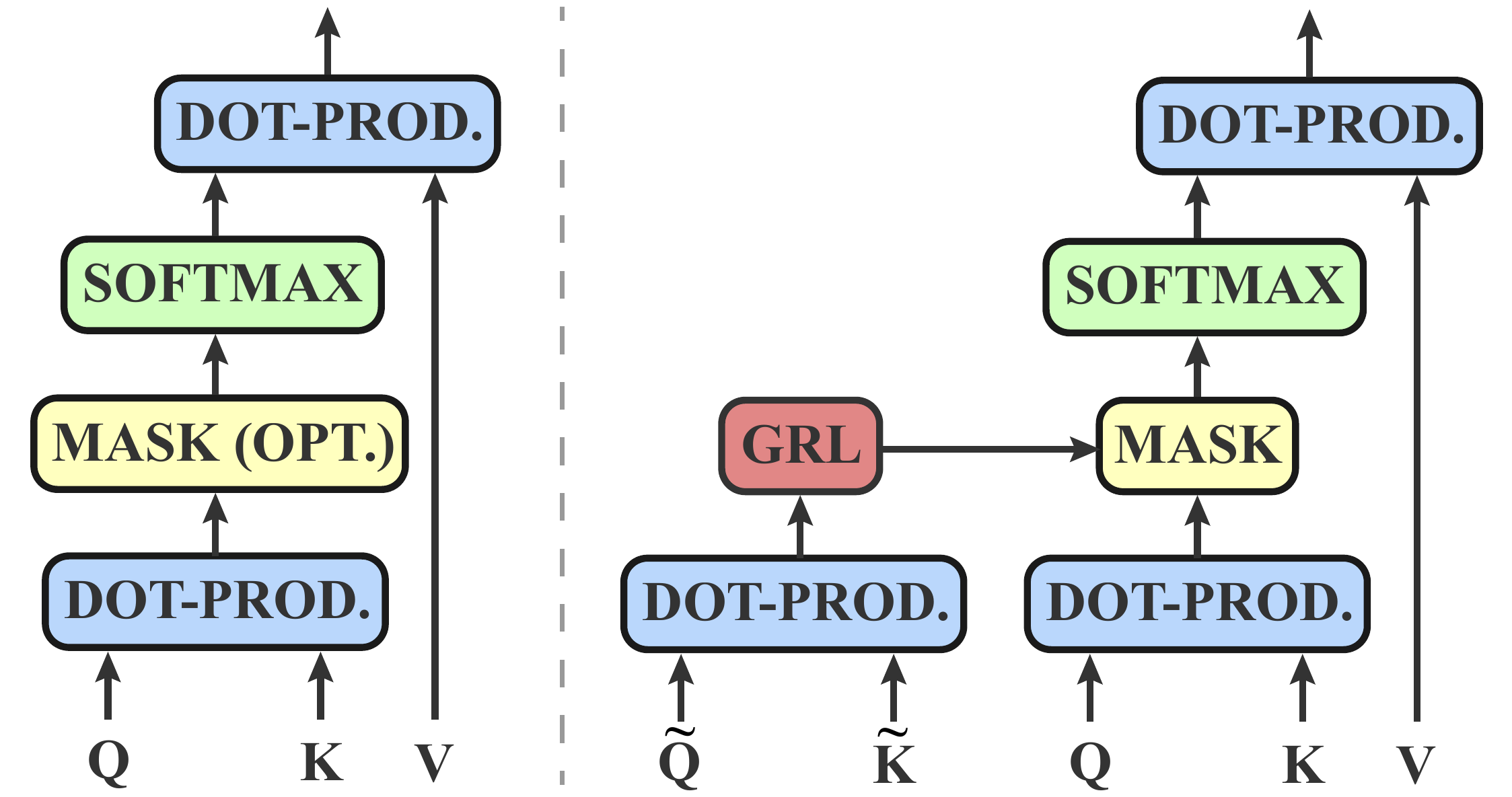}
\caption{Diagrams of a self-attention (SA) layer (left) and an ASA layer (right). Readers may refer to our supplementary material for the implementation detail.}
\label{f2}
\end{figure}

Adversarial training is naturally expensive. To remedy this, we propose a fast and simple implementation of ASA. There are two major points below.

\paragraph{Feature sharing} Adversarial training algorithms like $K$-PGD \citep{DBLP:conf/iclr/MadryMSTV18} barely avoid multiple inner optimization steps to achieve the high-quality solutions of the adversarial examples (for ASA, they are adversarial structures). Indeed, multiple inner steps could be a disaster for LMs, especially in the pre-training process, which will cost several days or weeks to go through once.

Though there are different ways to implement the ASA adversary $ \mu_{\mathbf x}=\mu(\mathbf x,\eta) $, we are supposed to allow it to obtain a nice solution of $ \mu_{\mathbf x}^* $ in few steps (e.g. only one step). Thus, for the $ i^{\rm th} $ self-attention layer, we let the input of $ \mu(\mathbf x,\eta)^i $ be the input hidden states $ \mathbf h^i $ of this layer. It does not contradict $ \mu(\mathbf x,\eta)^i $, since $ \mathbf h^i $ is encoded from $ \mathbf x $, suggesting that the adversary of each layer can access all useful features in the hidden states learned by the model from the lower layers. We apply two linear transformations on $ \mathbf h^i $ to obtain two components $ \widetilde Q $ and $ \widetilde K $, and take a dot-product of them to obtain the matrix $ {\widetilde Q \cdot \widetilde K^{T}}/{\sqrt{d}} $. It is a symmetrical process of computing $ {Q \cdot K^{T}}/{\sqrt{d}} $ in vanilla self-attention. The difference is that we will then binarize the matrix using the reparameterization trick \citep{DBLP:conf/iclr/JangGP17}.

Such a design allows us to take only one inner step but obtain a nice $ \mu_{\mathbf x}^* $. A potential risk is that we can not ensure the performance of ASA in the early training steps since $ \eta $ is randomly initialized. However, it is not naive since we generally utilize a very small learning rate at the beginning of LM training. The impact of these sacrificed training steps on the model can be negligible.

\paragraph{Gradient reversal} Another concern is that adversarial training algorithms always leverage an alternate optimization style, where we temporarily freeze one side of the model and the adversary, and update the other. It requires at least twice backward passes for the inner maximization and outer minimization. To further accelerate training, we adopt Gradient Reversal Layer (GRL) to merge two passes into one. GRL is first introduced in Domain-Adversarial Training \citep{DBLP:conf/icml/GaninL15}, acting as a switch during the backward pass to twist the sign of the gradient from the previous modules. The subsequent modules will be consequently optimized in the opposite direction.

We show a diagram in Figure \ref{f2}. We see that an ASA layer is composed of five components, where $ \widetilde Q $ and $ \widetilde K $ are competing with $ Q $ and $ K $ through GRL.

\subsection{Training}

We eventually present the training objective for training an ASA-empowered model. Given a task with labels, we let $ \mathcal L_e(\theta) $ be the task-specific loss (e.g. classification, regression). The model needs to make the right predictions against the ASA adversary so that we obtain the ASA loss $ \mathcal L_{asa}(\theta,\eta) = \mathcal D \left[p(y|\mathbf x,\theta),p(y|\mathbf x,\mu(\mathbf x,\eta),\theta)\right] $. At the same time, the adversary is subject to the penalty term $ \mathcal L_c(\eta) = \Vert \mu(\mathbf x,\eta) \Vert_1 $. The final training objective thus consists of three components:
\begin{equation}
\mathcal L_e(\theta) + \alpha \mathcal L_{asa}(\theta,\eta) + \tau \mathcal L_c(\eta)
\label{e9}
\end{equation}
where we find ASA performs just well when simply fixing the balancing coefficient $ \alpha $ to 1 so that $ \tau $ is the only hyperparameter of learning ASA.

Eq. \ref{e9} explains how ASA works. One point is that the penalty term is paralleled to the model since it is not associated with $ \theta $. We seek to find the optimal model parameters $ \theta $ to minimize the first two terms. On the other hand, because of GRL, we seek to find the optimal adversary parameters $ \eta $ to maximize the last two terms.

Though Eq. \ref{e9} covers all cases when fine-tuning the ASA-empowered model on downstream tasks. We will discuss more on pre-training. ASA is consistent with the current trend of self-supervised language modeling like MLM \citep{DBLP:conf/naacl/DevlinCLT19} and RTD \citep{DBLP:conf/iclr/ClarkLLM20}, where we construct the negative samples based on the super large corpus itself without additional labeling.

To be more concrete, we rely on MLM pre-training setting \citep{DBLP:conf/naacl/DevlinCLT19} in what follows and the other situations can be easily generalized. MLM intends to recover a number of masked pieces within the sequence and the loss of it is obtained from the cross-entropy on those selected positions, denoted as $ \mathcal L_{mlm} $. Thus, we can compute the divergence between the two model predictions before and after biased by ASA on those positions and obtain the token-level ASA loss $ \mathcal L^t_{asa}(\theta,\eta) $.

Aside from the masked positions, the beginning position is also crucial to PrLMs (e.g \texttt{[CLS]} in BERT), which is always used as an indicator for the relationship within the sequence (e.g. sentence order, sentiment). Thus, we pick this position out from the final hidden states and calculate the divergence on it as another part of the ASA loss $ \mathcal L^s_{asa}(\theta,\eta) $. Finally, pre-training with ASA can be formulated as:
\begin{equation}
\mathcal L_{mlm}(\theta) + \mathcal L^t_{asa}(\theta,\eta) + \mathcal L^s_{asa}(\theta,\eta) + \tau \mathcal L_c(\eta)
\label{e10}
\end{equation}
where $ \mathcal L^t_{asa} $ and $ \mathcal L^s_{asa} $ refer to the token-level and sentence-level ASA loss respectively.

Based on Eq. \ref{e10}, we may touch on the idea of ASA pre-training from two perspectives: (a) \textbf{Structural loss}: ASA acts as a regularizer on the empirical loss of language modeling (the same in Eq. \ref{e9}); (b) \textbf{Multiple objectives}: ASA can be viewed as two independent self-supervised pre-training objectives in addition to MLM.

\section{Experiments}
\label{s4}

Our implementations are based on \emph{transformers} \citep{wolf-etal-2020-transformers}.

\begin{table*}[t]{}
\centering
\setlength\tabcolsep{8.5pt}
\begin{tabular}{lccccccccccl}
\multicolumn{12}{c}{} \\ \toprule
\multirow{3}{*}{\textbf{Model}}
&  \multicolumn{3}{c}{\tiny \textit{\textbf{Sentiment Analysis / Inference}}}
&& \multicolumn{2}{c}{\tiny \textit{\textbf{Semantic Similarity}}}
&& \multicolumn{1}{c}{\tiny \textit{\textbf{NER}}}
&& \multicolumn{1}{c}{\tiny \textit{\textbf{MRC}}}
&  \multirow{3}{*}{\textbf{Avg}} \\
\cline{2-4} \cline{6-7} \cline{9-9} \cline{11-11} \rule{0pt}{11pt}
                    & SST-2                & MNLI                 & QNLI                 && QQP                  & STS-B                && WNUT-17              && DREAM                &                             \\
                    & (Acc)                & (Acc)                & (Acc)                && (F1)                 & (Spc)                && (F1)                 && (Acc)                &                             \\ \midrule
BERT                & 93.2                 & 84.1                 & 90.4                 && 71.4                 & 84.7                 && 47.8                 && 62.9                 & 76.4                        \\
\specialrule{-0.6em}{1pt}{1pt}
                    & \tiny $\pm$0.24      & \tiny $\pm$0.05      & \tiny $\pm$0.09      && \tiny $\pm$0.31      & \tiny $\pm$0.10      && \tiny $\pm$1.08      && \tiny $\pm$0.16      &                             \\
BERT$^{ASA}$        & \textbf{94.1}$\star$ & \textbf{85.0}        & \textbf{91.4}$\star$ && \textbf{72.3}        & \textbf{86.5}$\star$ && \textbf{49.8}$\star$ && \textbf{64.3}$\star$ & \textbf{77.6} $\uparrow$1.2 \\
\specialrule{-0.6em}{1pt}{1pt}
                    & \tiny $\pm$0.00      & \tiny $\pm$0.05      & \tiny $\pm$0.22      && \tiny $\pm$0.05      & \tiny $\pm$0.37      && \tiny $\pm$0.69      && \tiny $\pm$0.41      &                             \\ \midrule
BERT$\dagger$       & 93.5                 & 84.4                 & 90.5                 && 71.5                 & 85.4                 && 49.2                 && 61.2                 & 76.5 $\uparrow$0.1          \\
\specialrule{-0.6em}{1pt}{1pt}
                    & \tiny $\pm$0.32      & \tiny $\pm$0.08      & \tiny $\pm$0.13      && \tiny $\pm$0.08      & \tiny $\pm$0.36      && \tiny $\pm$0.94      && \tiny $\pm$0.82      &                             \\
BERT$^{ASA}\dagger$ & \textbf{94.0}        & \textbf{84.7}        & \textbf{91.5}$\star$ && \textbf{72.3}        & \textbf{86.5}$\star$ && \textbf{50.3}$\star$ && \textbf{63.3}$\star$ & \textbf{77.5} $\uparrow$1.1 \\
\specialrule{-0.6em}{1pt}{1pt}
                    & \tiny $\pm$0.05      & \tiny $\pm$0.06      & \tiny $\pm$0.19      && \tiny $\pm$0.05      & \tiny $\pm$0.23      && \tiny $\pm$0.55      && \tiny $\pm$0.28      &                             \\ \midrule
RoBERTa             & 95.6                 & 87.2                 & 92.8                 && 72.2                 & 88.4                 && 54.8                 && 67.0                 & 79.7                        \\
\specialrule{-0.6em}{1pt}{1pt}
                    & \tiny $\pm$0.05      & \tiny $\pm$0.15      & \tiny $\pm$0.21      && \tiny $\pm$0.05      & \tiny $\pm$0.17      && \tiny $\pm$0.80      && \tiny $\pm$0.62      &                             \\
RoBERTa$^{ASA}$     & \textbf{96.3}        & \textbf{88.0}        & \textbf{93.6}        && \textbf{73.7}$\star$ & \textbf{89.2}        && \textbf{57.3}$\star$ && \textbf{69.2}$\star$ & \textbf{81.0} $\uparrow$1.3 \\
\specialrule{-0.6em}{1pt}{1pt}
                    & \tiny $\pm$0.19      & \tiny $\pm$0.05      & \tiny $\pm$0.22      && \tiny $\pm$0.12      & \tiny $\pm$0.38      && \tiny $\pm$0.18      && \tiny $\pm$0.68      &                             \\ \bottomrule
\end{tabular}
\caption{Results on different tasks (mean and variance), where $\dagger$ refers to the longer-trained model with MLM. We run three seeds for GLUE sub-tasks (the first five, since only two test submissions are allowed each day) and five seeds for the others. For MNLI, we average the two scores of the ``m" and ``mm". $\star$ indicates the proposed approach unfolds $ >1 $ points absolute gain.}
\label{t1}
\end{table*}

\subsection{Setup}

We experiment on five NLP tasks down to 10 datasets:

$\bullet$ \textbf{Sentiment Analysis}: Stanford Sentiment Treebank (SST-2) \citep{DBLP:conf/emnlp/SocherPWCMNP13}, which is a single-sentence binary classification task;
$\bullet$ \textbf{Natural Language Inference (NLI)}: Multi-Genre Natural Language Inference (MNLI) \citep{DBLP:conf/naacl/WilliamsNB18} and Question Natural Language Inference (QNLI) \citep{DBLP:conf/iclr/WangSMHLB19}, where we need to predict the relations between two sentences;
$\bullet$ \textbf{Semantic Similarity}: Semantic Textual Similarity Benchmark (STS-B) \citep{DBLP:conf/semeval/CerDALS17} and Quora Question Pairs (QQP) \citep{DBLP:conf/iclr/WangSMHLB19}, where we need to predict how similar two sentences are;
$\bullet$ \textbf{Named Entity Recognition (NER)}: WNUT-2017 \citep{DBLP:conf/aclnut/AguilarMLS17}, which contains a large number of rare entities;
$\bullet$ \textbf{Machine Reading Comprehension (MRC)}: Dialogue-based Reading Comprehension (DREAM) \citep{DBLP:journals/tacl/SunYCYCC19}, where we need to choose the best answer from the three candidates given a question and a piece of dialogue;
$\bullet$ \textbf{Robustness learning}: Adversarial NLI (ANLI) \citep{DBLP:conf/acl/NieWDBWK20} for NLI, PAWS-QQP \citep{DBLP:conf/naacl/ZhangBH19} for semantic similarity, and HellaSWAG \citep{DBLP:conf/acl/ZellersHBFC19} for MRC.

We verify the gain of ASA on the top of two different self-attention (SA) designs: vanilla SA in BERT-base \citep{DBLP:conf/naacl/DevlinCLT19} and its stronger variant RoBERTa-base \citep{DBLP:journals/corr/abs-1907-11692}, and disentangled SA in DeBERTa-large \citep{DBLP:conf/iclr/HeLGC21}. In addition, we do the experiments on both pre-training ($ \tau = 0.1 $, Eq. \ref{e10}) and fine-tuning ($ \tau = 0.3 $, Eq. \ref{e9}) stages (the training details can be found in Appendix). For pre-training, we continue to pre-train BERT based on MLM with ASA on the English Wikipedia corpus. Besides, for fair enough comparison, we train another BERT with vanilla SA (BERT$\dagger$ in Table \ref{t1}). We set the batch size to 128 and train both models for 20K steps with FP16. Note that we directly fine-tune them without ASA.

\subsection{Results}

\begin{table}
\centering
\small
\begin{tabular}{@{}lccc@{}}
\toprule
\multirow{2}{*}{\textbf{Model}} & ANLI                          & PAWS-QQP                       & HellaSWAG               \\
                                & (Acc)                         & (Acc)                          & (Acc)                   \\ \midrule
BERT                            & 48.0$_{\pm.68}$               & 81.7$_{\pm1.24}$               & 39.7$_{\pm.28}$          \\
BERT$^{ASA}$                    & \textbf{50.4}$\star_{\pm.81}$ & \textbf{87.7}$\star_{\pm1.53}$ & \textbf{40.8}$\star_{\pm.27}$ \\ \midrule
DeBERTa (large)                 & 57.6$_{\pm.43}$               & 95.7$_{\pm.38}$                & 94.3$_{\pm1.02}$              \\
DeBERTa$^{ASA}$                 & \textbf{58.2}$_{\pm.94}$      & \textbf{96.0}$_{\pm.24}$       & \textbf{95.4}$_{\pm1.31}$      \\ \bottomrule
\end{tabular}
\caption{Results on robustness learning tasks when $ \tau = 0.3 $ over five runs. For ANLI, we put the test data of all rounds together and the model is trained with its own training data without any other data. For HellaSWAG, we report the dev.}
\label{t2}
\end{table}

\paragraph{Results on generalization} Table \ref{t1} summarizes the results across various tasks. For fine-tuning, ASA-empowered models consistently outweigh naive BERT and RoBERTa by a large margin, lifting the average performance of BERT from \textbf{76.4} to \textbf{77.6}, and RoBERTa from \textbf{79.7} to \textbf{81.0}. ASA is supposed to perform well on small sets like STS-B (84.7 to 86.5 on BERT) and DREAM (62.9 to 64.3) with merely thousands of training samples, which tend to be more susceptible to over-fitting. However, on much larger ones like MNLI (84.1 to 85.0), QNLI (90.4 to 91.4), and QQP (72.2 to 73.7 on RoBERTa) with more than hundred-thousands of samples, it still produces powerful gain. It implies that ASA not only enhances model generalization but also language representation. For continual pre-training, ASA brings competitive performance gain when directly fine-tuning on downstream tasks.

\paragraph{Results on robustness} To assess the impact of ASA on model robustness, we report the fine-tuning results on three challenging robustness benchmarks. These tasks contain a large number of adversarial samples in their training or test sets. From Table \ref{t2}, we can see that ASA produces 2.4, 6.0, and 1.1 points of absolute gain over BERT-base on the three tasks respectively. Even on strong DeBERTa-large, ASA can still deliver considerable improvement.

\section{Ablation Study}
\label{s5}

\subsection{VS. Naive Masking}

We compare ASA with three naive masking strategies. Bernoulli distribution is a widely-used priority in network dropout \citep{DBLP:journals/jmlr/SrivastavaHKSS14}. We report the best results with the masking probability selected in \{0.05, 0.1\}. Besides, we introduce another two potentially stronger strategies. For the first one, we dynamically schedule the masking probability for each step following the learned pattern by ASA. Different from ASA, the masked units here are Bernoulli chosen. For the second one, we always choose to mask those units with the most significant attention scores (a magnitude-based strategy). Similar to ASA, we apply the masking matrices to all self-attention layers, and the training objective corresponds to the first two terms of Eq. \ref{e9}.

From Table \ref{t5}, we can see that pure Bernoulli works the best among the three naive strategies, slightly better than scheduled Bernoulli. However, ASA outweighs them even by a large margin, suggesting that the worst-case masking can better facilitate model training. Besides, the magnitude-based masking turns out to be harmful. Since ASA acts as a gradient-based adversarial strategy, it may not always mask those globally most significant units in the attention matrix (this pattern can be seen in previous Figure \ref{f5}).

\begin{table}
\centering
\small
\begin{tabular}{@{}lccc@{}}
\toprule
                          & PAWS-QQP                    & HellaSWAG                     & WNUT-17                  \\ \midrule
\textit{Bernoulli}        & 86.1$_{\pm1.2}$             & 40.5$_{\pm0.2}$               & 48.2$_{\pm1.0}$          \\
\textit{Scheduled}        & 85.4$_{\pm1.5}$             & 40.2$_{\pm0.2}$               & 48.7$_{\pm0.9}$          \\
\textit{Magnitude}        & 84.6$_{\pm0.9}$             & 39.9$_{\pm0.3}$               & 47.3$_{\pm1.1}$          \\ \cdashline{0-3}[2pt/2pt]
\textit{ASA}              & \textbf{87.7}$_{\pm1.5}$    & \textbf{40.8}$_{\pm0.3}$      & \textbf{49.8}$_{\pm0.7}$ \\ \bottomrule
\end{tabular}
\caption{Naive masking on BERT-base over five runs.}
\label{t5}
\end{table}

On the other hand, it has been found in previous work that adversarial training on word embeddings can appear mediocre compared to random perturbations \citep{DBLP:conf/iclr/AghajanyanSGGZG21}. We are positive that adversarial training benefits model training, but hypothesize that the current spatial optimization of embedding perturbations suffers from shortcomings so that it sometimes falls behind random perturbations. However, the optimization of ASA is carefully designed in our paper.

\begin{table}
\centering
\small
\begin{tabular}{@{}lcccc@{}}
\toprule
                  & MNLI                     & QNLI                      & PAWS-QQP                     & HellaSWAG \\ \midrule
\textit{FreeLB}   & 85.3$_{\pm0.1}$          & 91.1$_{\pm0.0}$           & 86.3$_{\pm1.3}$              & 39.6$_{\pm0.4}$ \\
\textit{SMART}    & \textbf{85.5}$_{\pm0.2}$ & \textbf{91.6}$_{\pm0.5}$  & 85.8$_{\pm0.8}$              & 38.2$_{\pm0.3}$ \\
\textit{ASA}      & 85.0$_{\pm0.1}$          & 91.4$_{\pm0.2}$           & \textbf{87.7}$_{\pm1.5}$     & \textbf{40.8}$_{\pm0.3}$ \\ \bottomrule
\end{tabular}
\caption{Comparison with adversarial training on BERT-base over multiple runs (three for GLUE and five for others).}
\label{t6}
\end{table}

\subsection{VS. Adversarial Training}

ASA is close to conventional adversarial training, but there are two main differences. In the text domain, adversarial training works on the input space, imposing perturbations on word embeddings, while ASA works on model structures. Besides, adversarial training normally leverages projected gradient descent (PGD) \citep{DBLP:conf/iclr/MadryMSTV18} to learn the adversary, while ASA is optimized through an unconstrained manner. We compare the performances on different tasks between ASA and FreeLB \citep{DBLP:conf/iclr/ZhuCGSGL20}, one of the state-of-the-art adversarial training approaches in the text domain.

From Table \ref{t6}, we can see that ASA and FreeLB are competitive on MNLI and QNLI, while ASA outperforms by 1.4 and 1.2 points on PAWS-QQP and HellaSWAG. It may leave a new line for future research, where ASA can be superior to conventional adversarial training on certain tasks. Another advantage of ASA is that it only introduces one hyperparameter $ \tau $, while for FreeLB, we need to sweep through different adversarial step sizes and boundaries.

On the other hand, we find that both FreeLB and SMART can hardly induce a significant variation in the attention maps, even if the input embeddings are perturbed. The model remains focused on those pieces that are supposed to be focused on before being perturbed. This can be detrimental when one tries to explain the adversary's behavior.

\begin{figure}
\centering
\subfigure[Speed comparison]{
\includegraphics[width=0.22\textwidth]{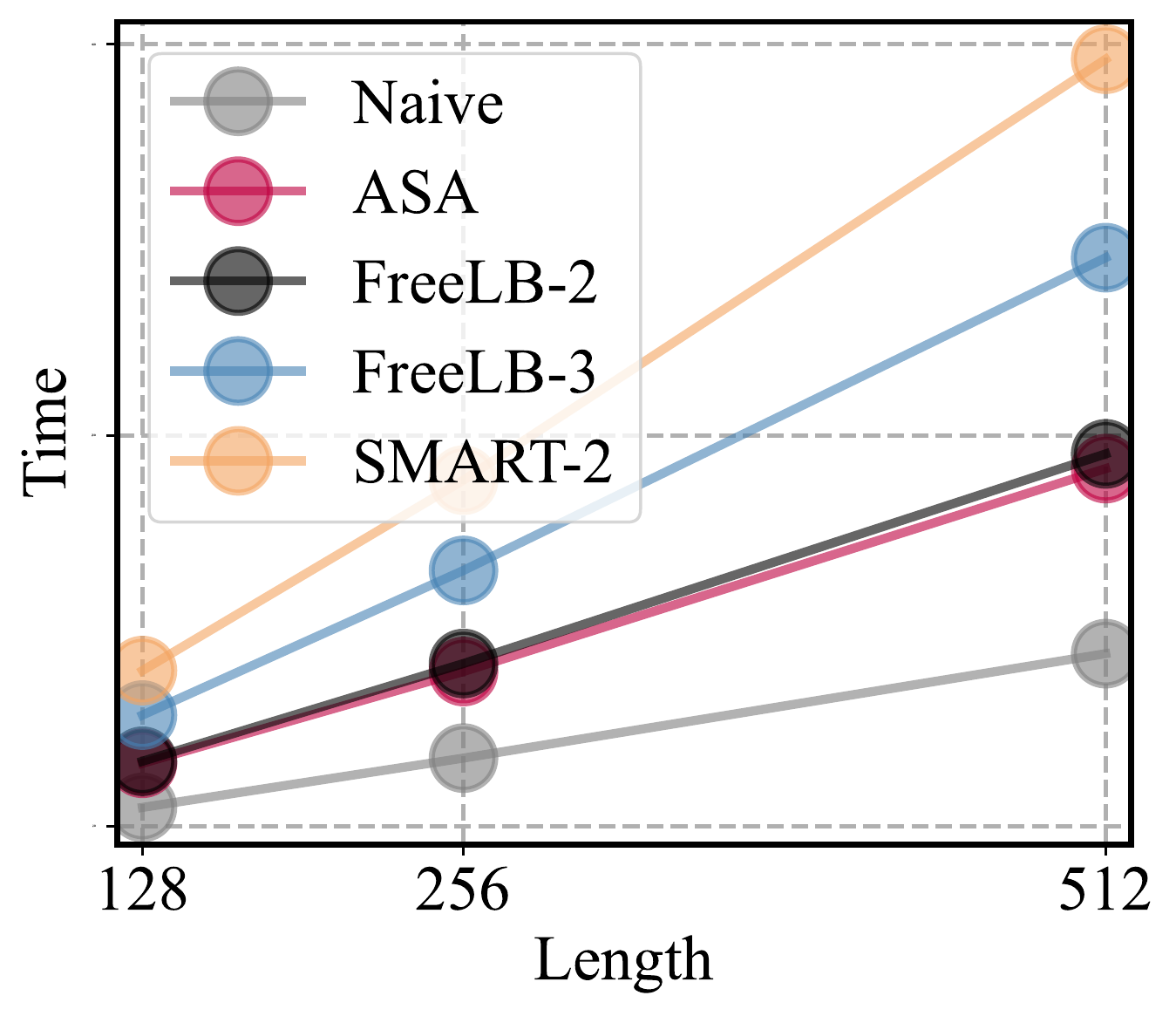}
}
\subfigure[Effect of temperature]{
\includegraphics[width=0.215\textwidth]{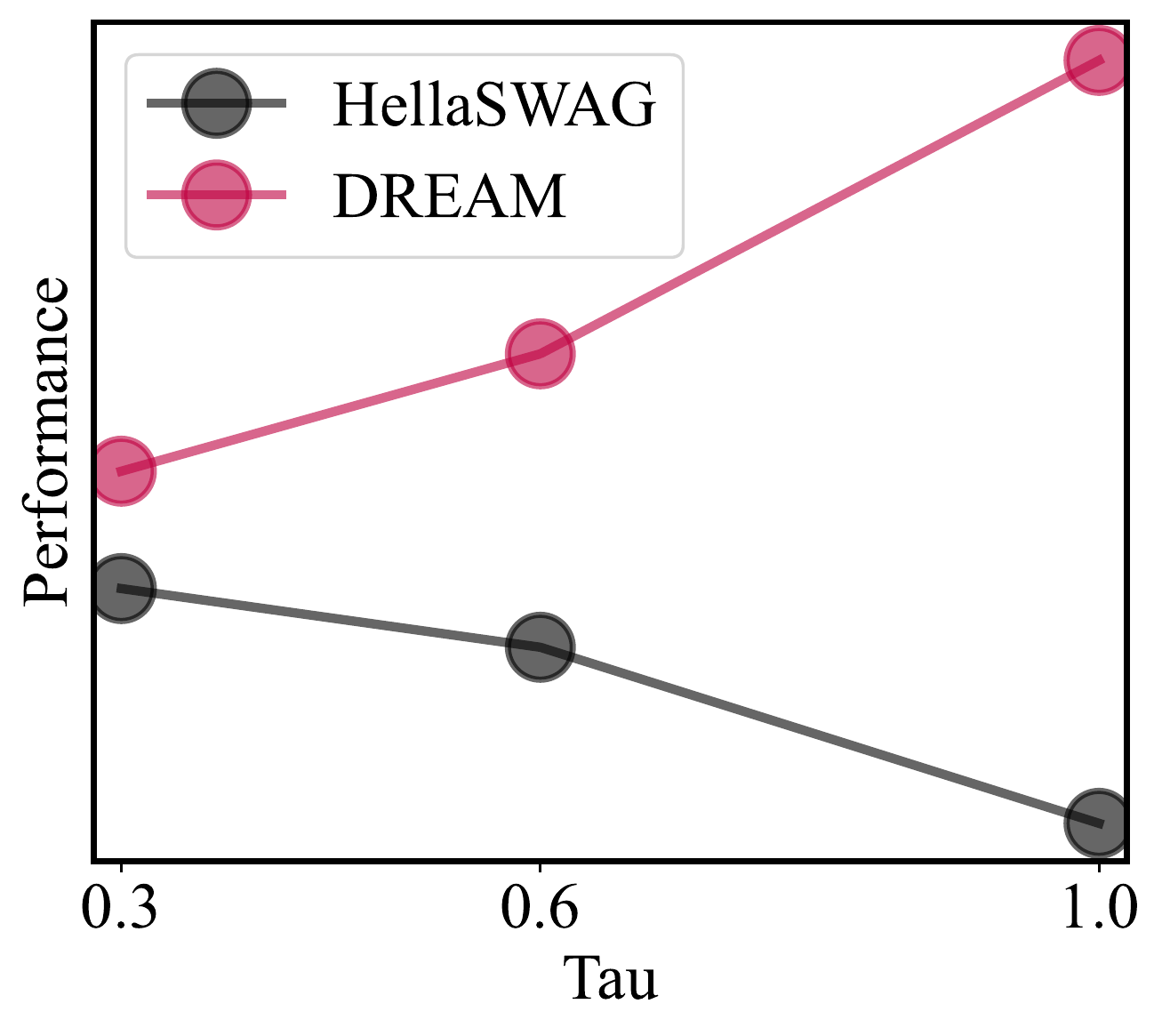}
}
\caption{Speed comparison of different learning approaches across sequence lengths (128, 256, and 512). FreeLB-$x$ means that we train the model with $x$ inner steps.}
\label{f4}
\end{figure}

\subsection{Training Speed}

Figure \ref{f4} (a) summarizes the speed performances of ASA and other two state-of-the-art adversarial training algorithms in the text domain. FreeLB \citep{DBLP:conf/iclr/ZhuCGSGL20} is currently the fastest algorithm, which requires at least two inner steps (FreeLB-2) to complete its learning process. SMART \citep{DBLP:conf/acl/JiangHCLGZ20} leverages the idea of virtual adversarial training, which thus requires at least one more forward passes. We turn off FP16, fix the batch size to 16, and do the experiments under different sequence lengths. We can see that ASA is slightly faster than FreeLB-2, taking about twice the period of naive training when the sequence length is up to 512. However, training with SMART and FreeLB-3 is much more expensive, taking about three and four times that period of naive training (SMART-1 is very close to FreeLB-3, so we omit it from the figure).

\subsection{Temperature Coefficient}

The only hyperparameter for ASA is the temperature coefficient $ \tau $, which controls the intensity of the adversary, a lower $ \tau $ corresponding to a stronger attack (higher masking proportion). In practice, $ \tau $ balances the model generalization and robustness. We conduct experiments on a benign task (DREAM) and an adversarial task (HellaSWAG) respectively with $ \tau $ selected in \{0.3, 0.6, 1.0\}. As in Figure \ref{f4} (b), the trends of two curves are opposite (we offset them vertically to make them close). A stronger adversary might lead to a decrease in generalization but benefit robustness. For example, we see the peak DREAM result when $ \tau=1.0 $, while the peak HellaSWAG result when $ \tau=0.3 $.

\subsection{Masking Proportion}

\begin{figure}
\centering
\subfigure[Between layers]{
\includegraphics[width=0.22\textwidth]{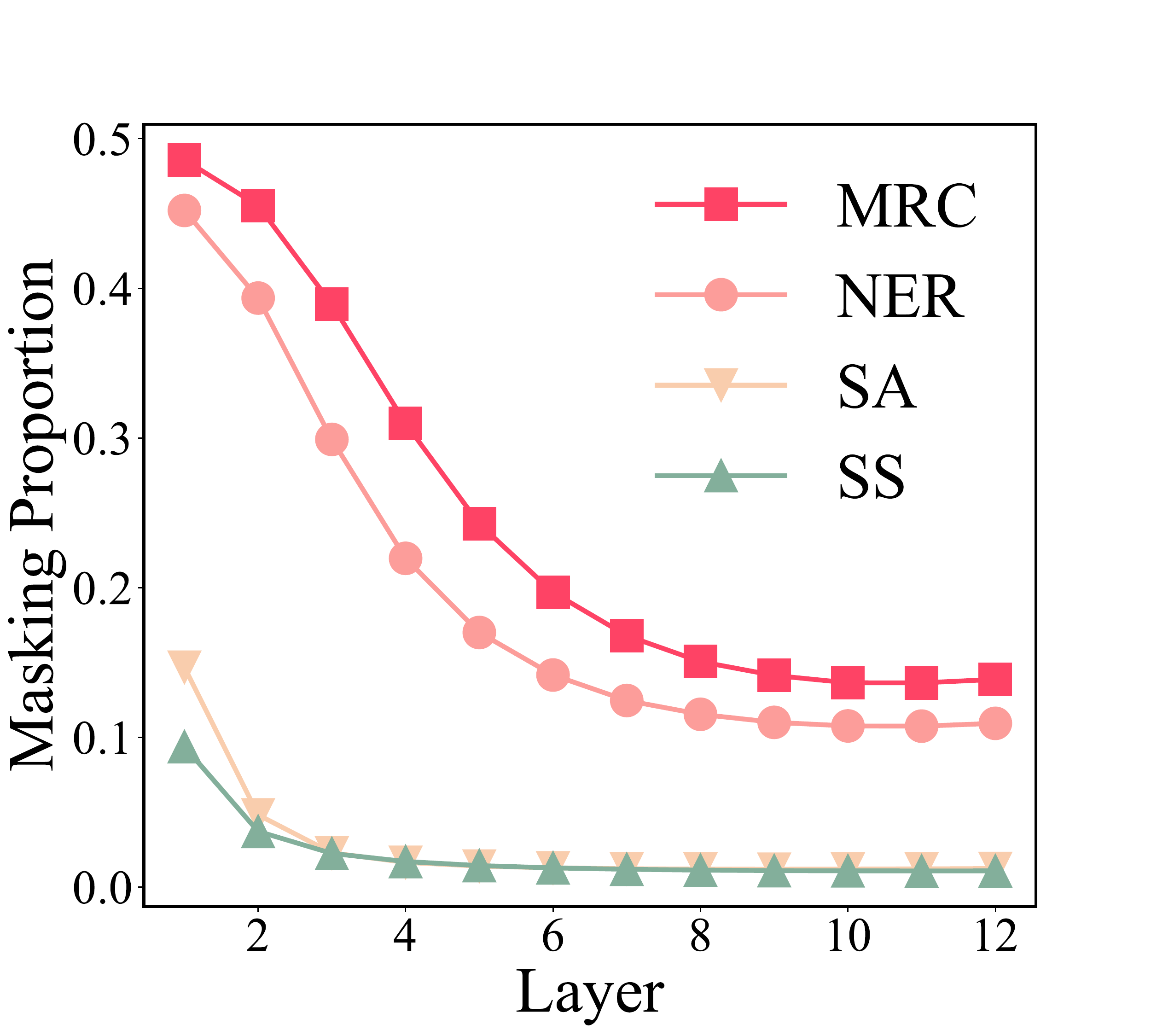}
}
\subfigure[Between tasks]{
\includegraphics[width=0.22\textwidth]{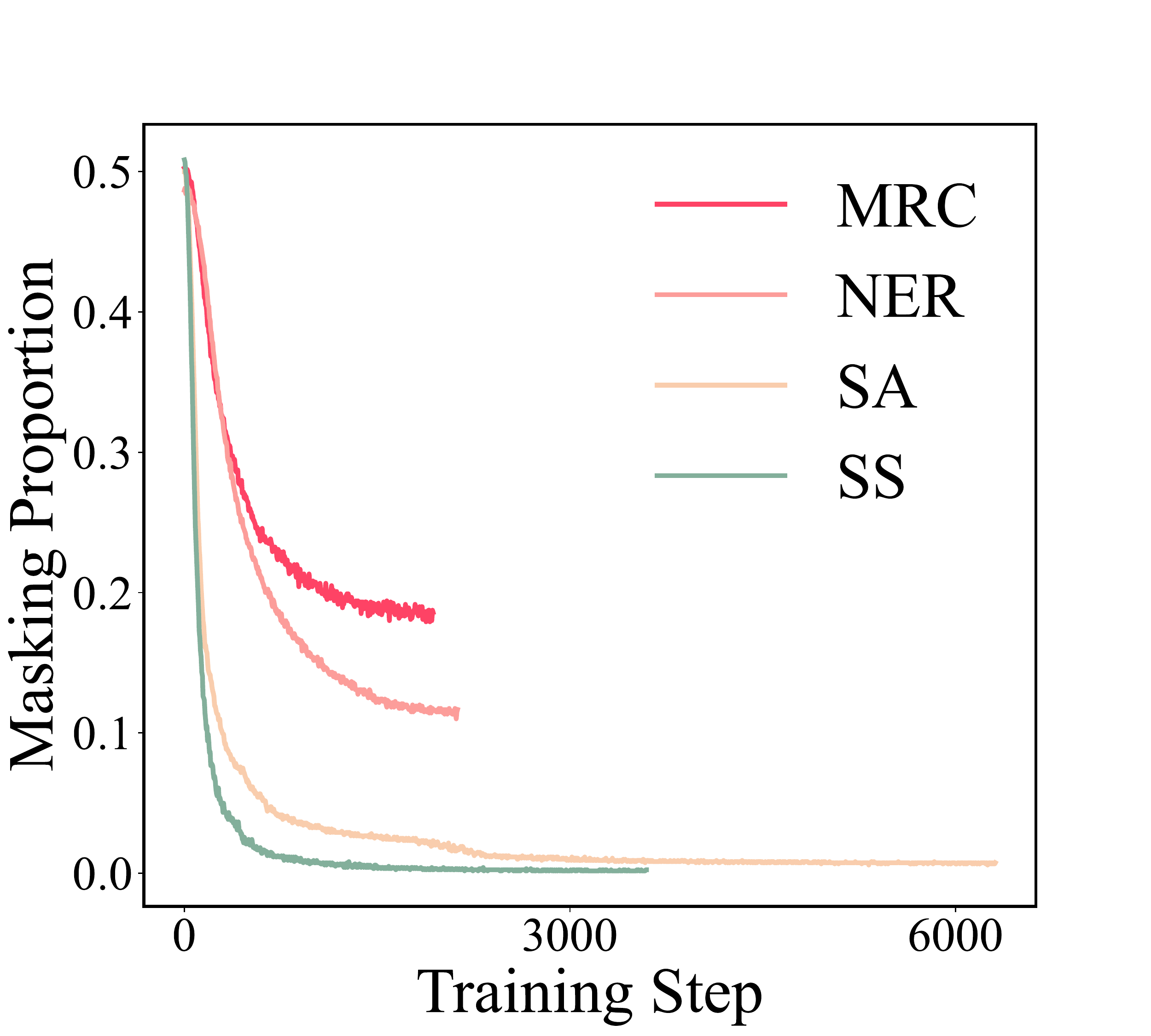}
}
\caption{Masking proportion of ASA when $ \tau = 0.3 $ across different tasks. Note that the magnitude of adversarial perturbations is always minor but they greatly fool the model.}
\label{f6}
\end{figure}

A higher masking proportion in ASA implies that the layer is more vulnerable. As in Figure \ref{f6} (a), we observe that the masking proportion almost decreases layer by layer from the bottom to the top. We attribute this to information diffusion \citep{DBLP:conf/icml/GoyalCRCSV20}, which states that the input vectors are progressively assimilating through continuously making self-attention. Consequently, the attention scores in the bottom layers are more important so that more vulnerable, while those in the top layers become less important after their assimilation. The feed-forward layers are more contributing this time \citep{DBLP:conf/acl/YouSI20}.

In Figure \ref{f6} (b), we calculate the average masking proportion of all layers and see that the situations can also be different between tasks even with the same temperature. Take sentiment analysis as an instance, the adversary merely needs to focus on specific keywords, which is enough to lead to misclassification. For NER and MRC, however, there are more sensitive words that are scattered across the sentence, and therefore a greater attack of the adversary is needed.

\section{Why ASA Works}

\begin{figure}
\centering
\subfigure[SA (wrong predicted)]{
\includegraphics[width=0.22\textwidth]{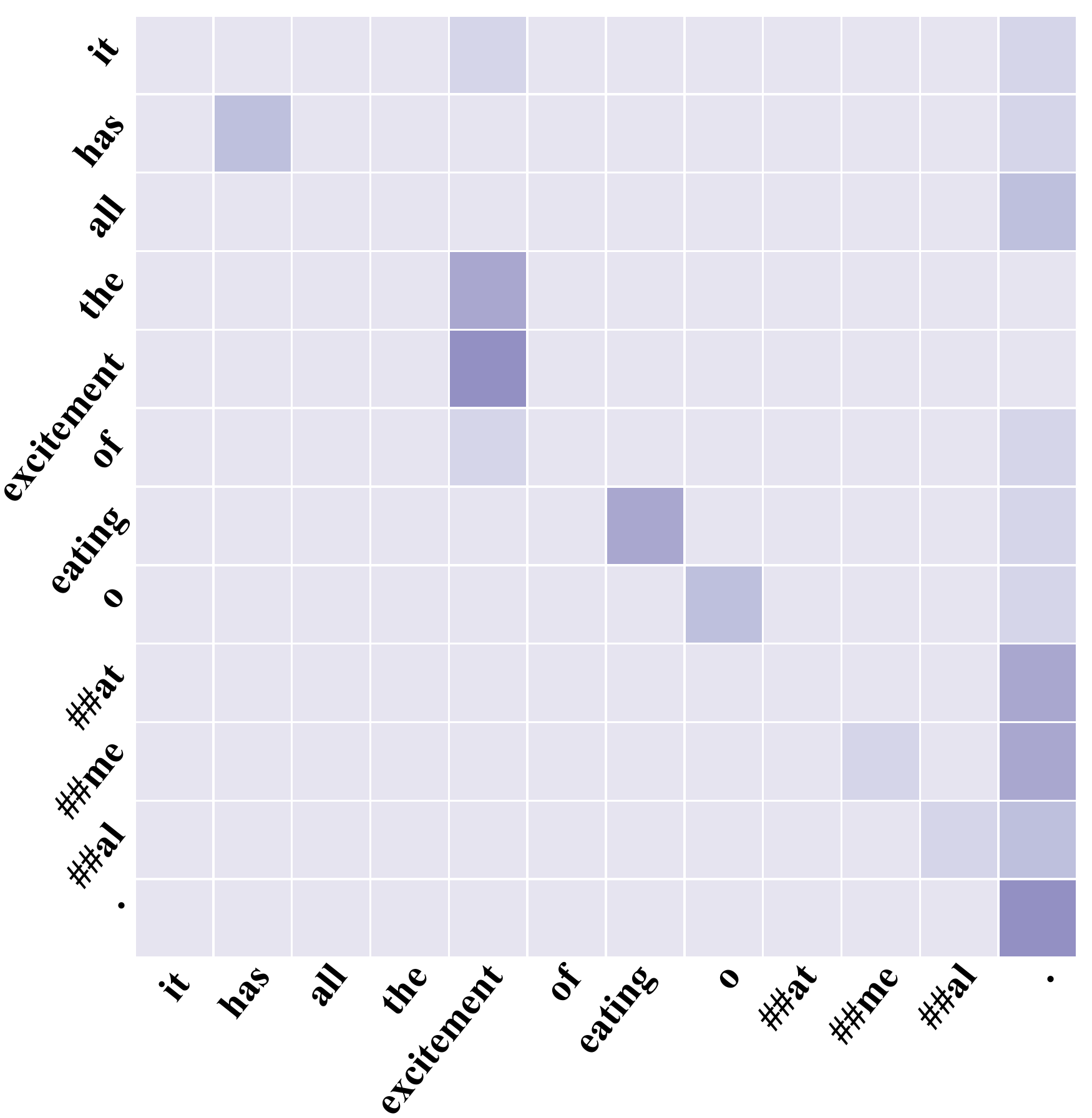}
}
\subfigure[ASA (correctly predicted)]{
\includegraphics[width=0.22\textwidth]{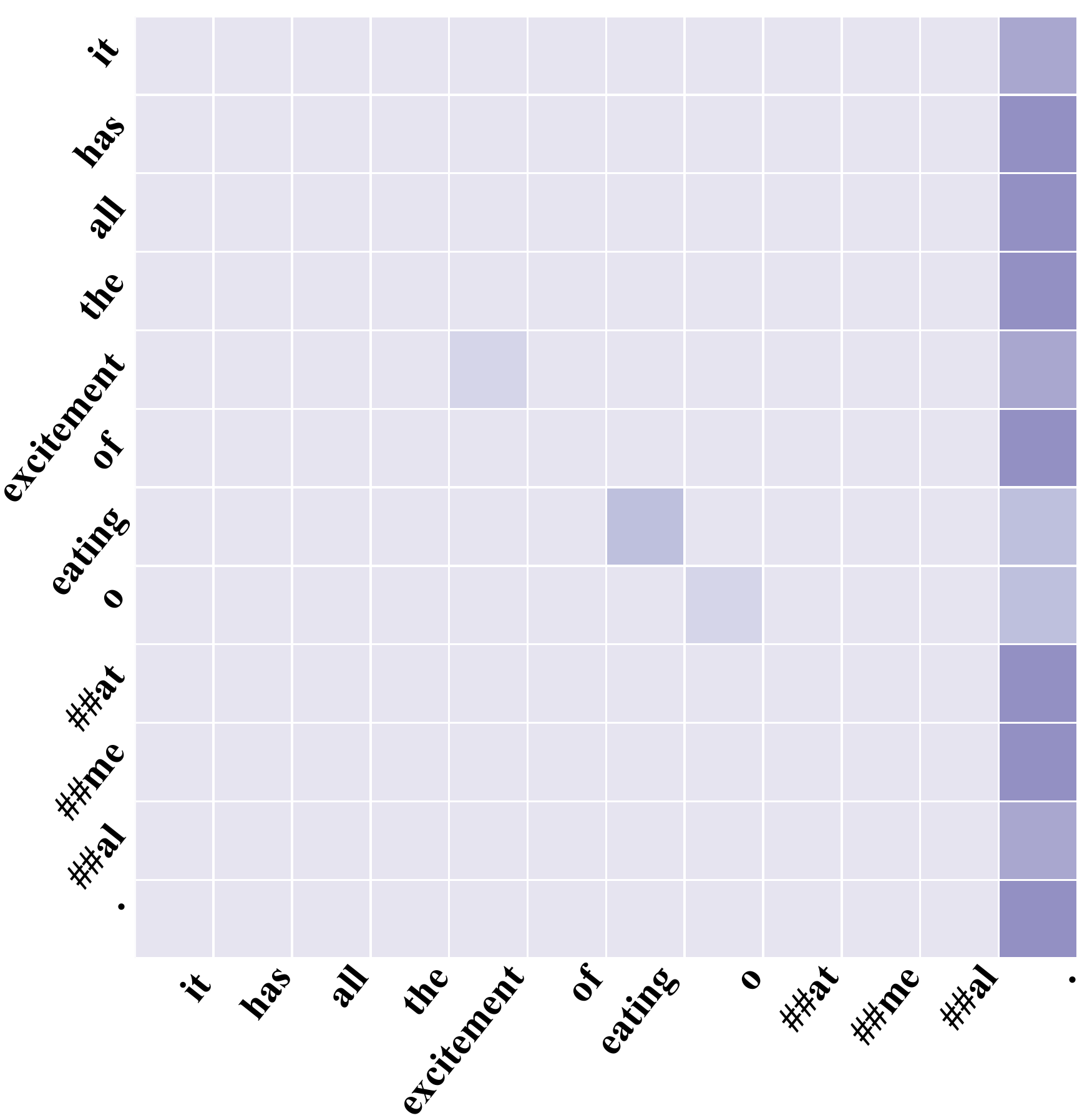}
}
\caption{Comparison of attention maps between SA-train and ASA-trained models. The text is \textit{it has all the excitement of eating oatmeal .} from SST-2). We choose the view from the 11$^{\rm th}$ layer and 2$^{\rm nd}$ head of BERT for instance.}
\label{f10}
\end{figure}

In addition, \textbf{ASA is effective in weakening the model reliance on keywords and encourages the model to concentrate on the broader semantics}. We show a concrete example of sentiment analysis in Figure \ref{f10}. We see that the SA-trained model tends to allow more tokens to receive strong attention (potential keywords), while they are sometimes spurious features. As a result, the model gives a wrong prediction. However, the ASA-trained one locates fewer keywords but more precisely (i.e. \textit{excitement}, \textit{eating}, \textit{oatmeal}) and thus obtains the right answer. Note that punctuation acts as a critical clue that signals the boundaries of semantics. In the top layers, they are normally given a high attention score.

Another observation is that for those examples without explicit keywords (not shown in the paper for space limitation), the SA-trained model prefers to locate many as ``keywords''. Contrarily, the ASA-trained one may not locate any keywords, but rather make the predictions based on entire semantics. The above observations are well-consistent with the way ASA trains.

\section{Related Work}

Our work is closely related to Adversarial Training (AT) \citep{DBLP:journals/corr/GoodfellowSS14}, which is a common machine learning approach to improve model robustness. In the text domain, the conventional philosophy is to impose adversarial perturbations on word embeddings \citep{DBLP:conf/iclr/MiyatoDG17}. It is later found to be highly effective in enhancing model performances when applied to fine-tuning on multiple downstream tasks, e.g. FreeLB \citep{DBLP:conf/iclr/ZhuCGSGL20}, SMART \citep{DBLP:conf/acl/JiangHCLGZ20}, InfoBERT \citep{DBLP:conf/iclr/WangWCGJLL21}, CreAT \citep{wu2023toward}, while ALUM \citep{DBLP:journals/corr/abs-2004-08994} provides the firsthand empirical results that adversarial training can produce a promising pre-training gain. As opposed to all these counterparts, which choose to perturb the input text or word embeddings, our work perturbs the self-attention structure. In addition, we present a new unconstrained optimization criterion to effectively learn the adversary. Our work is also related to smoothing and regularization techniques \citep{DBLP:journals/neco/Bishop95,DBLP:journals/jmlr/SrivastavaHKSS14}.

Adversarial training is naturally expensive. There are algorithms for acceleration, e.g. FreeAT \citep{DBLP:conf/nips/ShafahiNG0DSDTG19}, YOPO \citep{DBLP:conf/nips/ZhangZLZ019}, FreeLB \citep{DBLP:conf/iclr/ZhuCGSGL20}. This paper proposes a fast and simple implementation. Its speed performance rivals that of the current fastest adversarial training algorithm FreeLB. Another important line in adversarial training is to rationalize the behaviour of the adversary \citep{DBLP:conf/ijcai/SatoSS018}. In our work, we demonstrate how adversarial self-attention contributes to improving the model generalization from the perspective of feature utilization.

Our work is similar to optimizing the self-attention architecture \citep{DBLP:conf/nips/VaswaniSPUJGKP17}, e.g. block-wise attention \citep{DBLP:conf/nips/ZaheerGDAAOPRWY20}, sparse attention \citep{DBLP:conf/icml/ShiGRXLLK21}, structure-induced attention \citep{DBLP:conf/acl/WuZZ21}, Gaussian attention \citep{DBLP:conf/acl/YouSI20}, synthetic attention \citep{DBLP:conf/icml/TayBMJZZ21}, policy-based attention \citep{DBLP:journals/corr/abs-2104-04692}. Most of these variants are based on a predefined priority. In comparison, our adversary derives from the data distribution itself and exploits the adversarial idea to effectively learn the self-attention structure or how to make self-attention. It is a kind of Meta-Learning \citep{DBLP:books/sp/98/ThrunP98}, which aims to effectively optimize the learning process, e.g. learning the update rule for few-shot learning \citep{DBLP:conf/iclr/RaviL17}, learning an optimized initialization ready for fast adaption to new tasks \citep{DBLP:conf/icml/FinnAL17}, reweighting training samples \citep{DBLP:conf/icml/RenZYU18}. Our work facilitates both fine-tuning and pre-training for pre-trained language models (PrLMs) \citep{DBLP:conf/naacl/DevlinCLT19,DBLP:journals/corr/abs-1907-11692,DBLP:journals/jmlr/RaffelSRLNMZLL20,DBLP:conf/iclr/HeLGC21}. It is agnostic to the current pre-training paradigms, e.g. MLM \citep{DBLP:conf/naacl/DevlinCLT19}, RTD \citep{DBLP:conf/iclr/ClarkLLM20}, PLM \citep{DBLP:conf/nips/YangDYCSL19}, and multiple objectives \citep{DBLP:journals/corr/abs-2210-10293}.

\section{Conclusion}

This paper presents \textit{Adversarial Self-Attention} mechanism (ASA) to improve pre-trained language models. Our idea is to adversarially bias the Transformer attentions and facilitate model training from contaminated model structures. As it turns out, the model is encouraged to explore more on broader semantics and exploit less on keywords. Empirical experiments on a wide range of natural language processing (NLP) tasks demonstrate that our approach remarkably boosts model performances on both pre-training and fine-tuning stages. We also conduct a visual analysis to interpret how ASA works. However, the analysis in this paper is still limited. Future work can further dissect the Transformer attentions on more complicated tasks (e.g. MRC, reasoning).

\bibliography{aaai23}

\appendix

\section{Training Details}

\begin{table}[H]
\centering
\small
\begin{tabular}{@{}lcccccc@{}}
\toprule
          & LR     & BSZ     & EP      & WP      & MSL     & TMP \\ \midrule
SST-2     & 2e-5   & 32      & 3       & 0.06    & 128     & 0.3 \\
MNLI      & 3e-5   & 32      & 3       & 0.06    & 128     & 0.3 \\
QNLI      & 2e-5   & 32      & 3       & 0.06    & 128     & 0.3 \\
QQP       & 5e-5   & 32      & 3       & 0.06    & 128     & 0.3 \\
STS-B     & 5e-5   & 16      & 3       & 0.06    & 128     & 0.3 \\
WNUT-17   & 5e-5   & 16      & 5       & 0.1     & 64      & 0.5 \\
DREAM     & 3e-5   & 16      & 8       & 0.1     & 128     & 0.5 \\
ANLI      & 3e-5   & 32      & 3       & 0.06    & 128     & 0.3 \\
PAWS-QQP  & 5e-5   & 16      & 3       & 0.06    & 128     & 0.3 \\
HellaSWAG & 2e-5   & 32      & 3       & 0.1     & 128     & 0.3 \\ \bottomrule
\end{tabular}
\caption{Suggested fine-tuning setting. LR: learning rate; BSZ: batch size; EP: training epochs; WP: warmup proportion; MSL: sequence length; TMP: temperature coefficient.}
\label{t7}
\end{table}

\begin{table}[H]
\centering
\small
\begin{tabular}{@{}lcc@{}}
\toprule
                          & Adversarial   & Regular        \\ \midrule
TMP                       & 0.1           & -              \\
Dropout                   & 0.1           & 0.1            \\
Batch size                & 128 * 8       & 128 * 8        \\
Learning rate             & 2e-5          & 2e-5           \\
Weight Decay              & 0.01          & 0.01           \\
Max sequence length       & 256           & 256            \\
Warmup proportion         & 0.06          & 0.06           \\
Max steps                 & 20K           & 20K            \\
Gradient clipping         & 1.0           & 1.0            \\
FP16                      & Yes           & Yes            \\
Number of GPUs            & 8             & 8              \\ \bottomrule
\end{tabular}
\caption{Suggested pre-training setting.}
\end{table}

\end{document}